\documentclass{article}
\usepackage{multicol}
\usepackage{wrapfig}
\usepackage{times}
\usepackage{amsmath}
\usepackage{amssymb}
\usepackage{siunitx}
\usepackage{gensymb}
\usepackage{booktabs}
\usepackage{multirow}
\usepackage{authblk}
\usepackage{pifont}
\usepackage{soul}
\usepackage[symbol]{footmisc}

\usepackage{mdframed}
\usepackage{makecell}
\usepackage{soul}
\usepackage{todonotes}
\usepackage{subcaption}
\usepackage{xcolor}
\usepackage{adjustbox}
\usepackage{pdflscape}
\usepackage{bbm}
\usepackage{anyfontsize}

\usepackage{cite} 

\usepackage{hyperref}
\hypersetup{
    colorlinks=true,
    linkcolor=blue,
    urlcolor=blue,
}
\usepackage[moderate]{savetrees}
\usepackage{enumitem}
\setlist[itemize]{leftmargin=2em}
\setlist[enumerate]{leftmargin=2em}

\definecolor{mygreen}{rgb}{0, 0.5, 0}
\definecolor{mygrey}{rgb}{0.6, 0.6, 0.6}

{\list{}{\leftmargin=0.3in\rightmargin=0.3in}\item[]}%
{\endlist}

\newcommand{\mysubsection}[1]{\vspace{-2mm}\subsection{#1}\vspace{-2mm}}

\definecolor{myPink}{rgb}{0.87, 0.36, 0.51}

\definecolor{light-gray}{gray}{0.95}

\definecolor{dark-gray}{HTML}{A9A9A9} 
\definecolor{light-gray}{HTML}{b7b7b7} 
\definecolor{bg-gray}{HTML}{F8F8F8} 
	
\definecolor{green}{HTML}{6aa84f} 
\definecolor{highlight}{HTML}{cfe2f3} 
\definecolor{blue}{HTML}{0000ff} 
\definecolor{brown}{HTML}{9F8C76}
\definecolor{pink}{HTML}{D88782}
\definecolor{light-blue}{HTML}{4a86e8}
\definecolor{dark-yellow}{HTML}{F6AE2D}

\usepackage[preprint]{corl_2026} 

\usepackage{xspace}
\def\OURSFULL{Universal Manipulation Exoskeleton\xspace}
\def\OURS{UME\xspace}
\newcount\Comments 
\usepackage{color}
\definecolor{darkgreen}{rgb}{0,0.5,0}
\definecolor{purple}{rgb}{1,0,1}
\newcommand{\kibitz}[2]{\ifnum\Comments=1\textcolor{#1}{#2}\fi}

\usepackage{float}
\usepackage{hyperref}

\title{Universal Manipulation Exoskeleton: Learning Compliant Whole-body Policies with Real-time Torque Feedback}

\author{
    \vspace{0.8mm}\textbf{Litian Liang}\textsuperscript{*,1} ~~ \textbf{Jingxi Xu}\textsuperscript{*,1,2} ~~ \textbf{Xinda Qi}\textsuperscript{1} ~~ \textbf{Yujun Cai}\textsuperscript{1} ~~ \textbf{Houzhu Ding}\textsuperscript{1} ~~ \textbf{Luqi Wang}\textsuperscript{1}\\
    \vspace{2mm}
    \textup{\normalsize \textbf{Zhixin Sun}\textsuperscript{1} ~~ \textbf{Jyh-Herng Chow}\textsuperscript{1} ~~ \textbf{Ming Yang}\textsuperscript{1} ~~ \textbf{Mark Cutkosky}\textsuperscript{2}} \\
    \vspace{0.8mm}
    \normalsize \textup{\textsuperscript{1}Ant Group ~~ \textsuperscript{2}Stanford University}\\
    \vspace{2mm}
    \normalsize \textup{\textsuperscript{*}Equal contribution}\\
    \textup{\url{https://ume-exo.github.io/}}
    \vspace{-5mm}
}

\begin{document}
\maketitle


\begin{figure}[h] 
    \vspace{-4mm}
    \centering
    \includegraphics[width=\linewidth]{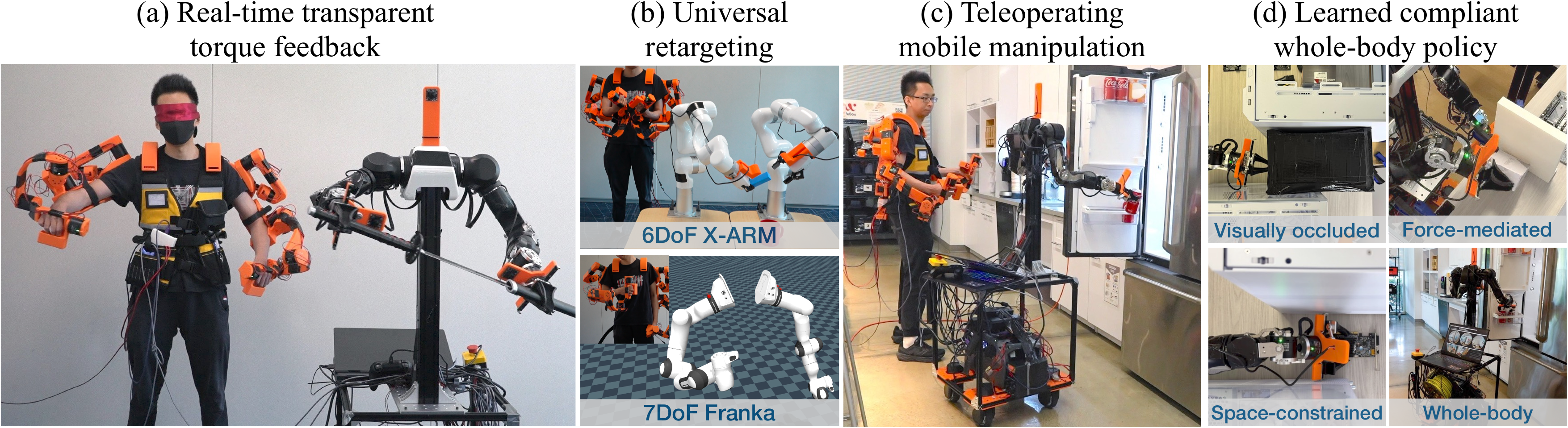}
    \caption{\footnotesize Overview of \OURSFULL. (a) \OURS provides transparent haptic torque feedback in real time, enabling the operator to teleoperate a bimanual robot to unsheathe a sword while blindfolded. (b) Our universal retargeting algorithm enables intuitive teleoperation across a range of robotic arms, including OpenArm (7-DoF), Franka (7-DoF), and X-ARM (6-DoF). (c) \OURS is portable and incorporates embedded IMU sensors that allow the user to simultaneously control the mobile base, enabling mobile manipulation teleoperation. (d) The robot learns whole-body compliant behaviors for highly force-mediated and extremely constrained-space tasks.}
    \label{fig:teaser}
    \vspace{-2mm}
\end{figure}

\begin{abstract}
    For robots to work safely in household environments, they need to be compliant and react to torque and force feedback during contact. However, the majority of existing data collection pipelines still lack the ability to capture force and torque data for learning active compliant policies.
    In this paper, we present Universal Manipulation Exoskeleton (\OURS), an upper-limb exoskeleton that provides real-time haptic torque feedback while recording whole-arm configurations and joint torque signals for teleoperation. With transparent torque feedback, human operators can even unsheathe kinematically constrained objects while blindfolded.
    \OURS is low-cost, lightweight, and portable. Equipped with an embedded IMU, it enables teleoperation for mobile manipulation. With our proposed universal retargeting algorithm, \OURS can teleoperate a range of robots, including the 7DoF OpenArm, 7DoF Franka, and 6DoF X-ARM. We demonstrate that this combination of capabilities enables learning bimanual, whole-body, and active compliant policies that operate effectively in highly constrained spaces. 
    The learned robust autonomous policies achieve high success rates across a variety of tasks, including long-horizon mobile manipulation, force-mediated box flipping, visually occluded box pushing, and space-constrained tabletop manipulation.
\end{abstract}



\section{Introduction and Related Work}



Industrial robots have advanced tremendously over the past few decades, becoming precise, fast, and rigid; however, if you get in their way, their strength and rigidity can pose serious safety risks. With the growing interest in collaborative, household robots, active compliance is becoming increasingly important, as these systems must sense and adapt to physical interactions rather than execute rigid motions. Nonetheless, much of the recent work on vision-language-action (VLA) models~\citep{black2024pi0,black2025pi05,pi2025pi06,black2026pi07,barreiros2026careful,wu2026pragmatic} and world models for robotics~\citep{wang2026interactive,bruce2024genie,chen2025large,li2025novaflow} does not emphasize compliance, largely due to the lack of force and torque data. While recent efforts have begun incorporating force and tactile information into robotic foundation models~\citep{xie2025towards} and adaptive data collection pipelines~\citep{xu2026compliant}, learning compliant whole-body manipulation at scale remains an open challenge.

Teleoperation is an important paradigm for collecting human demonstrations for learning active compliant policies. However, systems such as ALOHA~\cite{Zhaoetal2023,Fuetal2024} and GELLO~\cite{Wu2024GELLO} do not expose joint torque data during data collection. This leads to two key limitations. First, without torque data, robots cannot learn active compliant behaviors. Second, the lack of torque sensing prevents propagating resistance back to the leader arm, meaning that operators cannot feel the forces experienced by the follower arm. This absence of real-time haptic torque feedback limits both intuitiveness and safety, especially in force-intensive tasks. For example, in non-prehensile manipulation where a robot leverages environmental contact (e.g., using wall resistance to flip a box, shown in Fig.~\ref{fig:teaser}(d)), the absence of feedback may cause users to inadvertently overdrive the leader arm, which can become unsafe when the resistance on the follower arm suddenly disappears. In addition, these systems typically provide control only at the end-effector level, without direct control over the whole-arm configuration. However, whole-arm coordination can be critical for executing tasks in highly constrained spaces.


Another paradigm is hand-held data collection, such as UMI-style grippers~\citep{Chietal2024,Xuetal2025DexUMI,choi2026inthewildcompliantmanipulationumift}, which do provide direct force and torque feedback to the user during data collection, arising naturally from in-the-wild physical interactions. 
However, despite this implicit haptic feedback, the UMI-style paradigm typically records only end-effector poses and thus relies on inverse kinematics (IK) to enforce collision constraints at inference time. While this design promotes cross-embodiment generalization, it does not handle cluttered or highly constrained environments well. 
In addition, because the camera is often mounted on the gripper, the resulting observations may have limited coverage of the global scene geometry, which is critical for reliable collision avoidance, though recent works such as HoMMI~\citep{xu2026hommi} mitigate this issue using head-mounted cameras.



To address some aforementioned challenges, we present \OURS (\textbf{U}niversal \textbf{M}anipulation \textbf{E}xoskeleton). The key design principle of \OURS is to enable the teleoperator to directly feel, in real time, the resistance sensed by the robot. This bidirectional interaction provides a more intuitive teleoperation experience. Moreover, the recorded joint-level torque signals can be leveraged to train compliant whole-body and bimanual policies, especially in highly space-constrained environments.

Humans naturally regulate interaction forces during contact-rich manipulation. When retrieving a bottle from the back of a cluttered refrigerator, incidental contact with surrounding objects is common. Rather than strictly avoiding these contacts, humans adaptively modulate force to accomplish the task safely. Traditionally, robotic systems are designed to avoid contact; in contrast, our approach exploits torque feedback to learn behaviors that deliberately utilize contact. With the haptic torque feedback, it naturally captures human strategies for active compliance; with the recorded whole-arm joint torques and positions, it enables policies to implicitly encode aspects of motion planning from visual inputs and learn such contact-rich behavior. 

Recent advances in low-reduction, high-torque-density actuators are making it feasible to build a low-cost, lightweight, portable exoskeleton capable of delivering transparent haptic torque feedback. Compared to the Dynamixel~\citep{dynamixel} actuators used in prior works~\cite{aldaco2024aloha, Wu2024GELLO}, which typically rely on multi-stage planetary gearboxes with high gear ratios, we instead adopt quasi-direct-drive motors with substantially lower reduction ratios (such as Damiao~\citep{damiao} and Unitree~\citep{unitree}). \OURS is also equipped with an onboard inertial measurement unit (IMU), enabling teleoperation of dual-arm mobile manipulators for whole-body tasks in dynamic environments. 





In summary, \OURS is an upper-limb teleoperation exoskeleton with the following capabilities.

\begin{itemize}[leftmargin=*, itemsep=0pt, topsep=0pt]
\label{bullet}
\item \label{item:cap1} \textbf{Cap1.~Real-time haptic torque feedback.} \OURS provides real-time haptic torque feedback to the teleoperator, allowing them to feel the transparent resistance sensed by the robot. 
\item \label{item:cap2} \textbf{Cap2.~Universal whole-arm configuration capture.} \OURS records whole-arm configuration, and with our proposed retargeting algorithm, it can universally teleoperate a range of robots.
\item \label{item:cap3} \textbf{Cap3.~Portability for mobile manipulation.} \OURS is low-cost, lightweight, and portable. The embedded IMU sensor enables the teleoperation of mobile manipulation tasks.
\end{itemize}

To the best of our knowledge, we are the first to demonstrate that this combination of capabilities enables the learning of whole-body, bimanual, and active compliant policies, which significantly improve success rates on highly force-mediated tasks and tasks performed in tightly constrained spaces. 
For extended related work on teleoperated and in-the-wild data collection, and learning active compliant behaviors, see Appendix~\ref{sec:related_work}.

\section{Method}

We made several key design choices to achieve transparent haptic torque feedback. \OURS closely matches the human arm kinematic structure, improving wearing comfort and enabling universal teleoperation across robot arms with different DoFs (degrees of freedom). To support mobile manipulation teleoperation, we additionally use a low-cost IMU to control the mobile base through the operator’s upper-body orientation.


\mysubsection{Coaxial Actuator Configuration}

\begin{figure}[h]
    \centering
    \includegraphics[width=\linewidth]{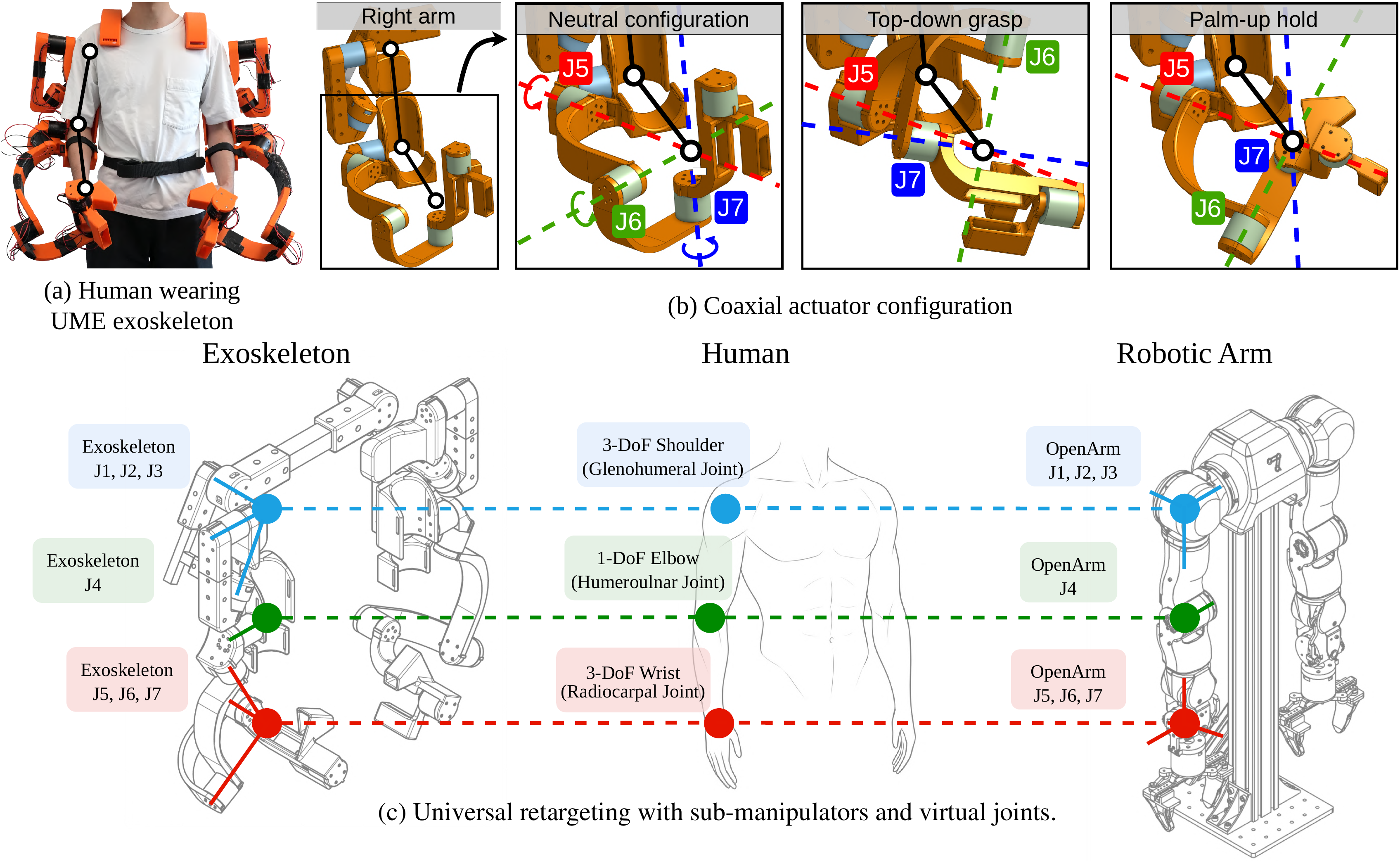}
    \vspace{-8mm}
    \caption{\footnotesize \OURS mechanical design.}
    \label{fig:mechanical}
    \vspace{-2mm}
\end{figure}

Our exoskeleton prioritizes comfort for the human arm. The human arm follows a 3-1-3 structure~\cite{perry2007upper,gopura2009mechanical}. The first three DoFs correspond to the shoulder (glenohumeral) joint, a ball-and-socket joint with three rotational DoFs: flexion/extension, abduction/adduction, and internal/external rotation. The elbow (humeroulnar) joint is a hinge joint with one DoF: flexion/extension. The final three DoFs correspond to the wrist (radiocarpal) joint, which provides flexion/extension, radial/ulnar deviation, and pronation/supination.

Inspired by~\citet{zimmermann2019anyexo,zimmermann2023anyexo}, we arranged the actuators in a coaxial configuration, where the center axes of joints J1, J2, and J3 intersect at the center of the human’s shoulder ball joint.
This ensures that, regardless of how the joint angles of J1, J2, and J3 change, their axes always intersect, even as the angles between the axes dynamically vary (shown in Fig.~\ref{fig:mechanical}(b)). Joint J4 corresponds to the human elbow joint. Finally, joints J5, J6, and J7 are arranged in another coaxial configuration, whose axes intersect at the center of the human wrist ball joint. Unlike other isomorphic leader-follower teleoperation systems that often have structural links blocking human movements, our special mechanical design replicates the \emph{full} range of motion (RoM) of the human wrist and the majority of the shoulder (see Fig.~\ref{fig:range_of_motion} in Appendix~\ref{appendix:range_of_motion} for a visualization of \OURS RoM).

\mysubsection{Universal Retargeting}



Our retargeting framework is robot-agnostic. 
We use the OpenArm system~\cite{damiao,openarm2025} as an example to demonstrate its efficacy (Fig.~\ref{fig:mechanical}(c)).
Inspired by the decoupling philosophies in Macro-Mini manipulator architectures \cite{khatib1991reduced}, we decompose the kinematic chain of both the exoskeleton (\OURS) and the follower robot into three sub-manipulators: a virtual 3-DoF spherical shoulder joint, a 1-DoF elbow joint, and a virtual 3-DoF spherical wrist joint. Motion and haptic torques are retargeted independently across these sub-structures, maximizing system stability, predictability, and intuitiveness.


\textbf{Position and Velocity Feedforward.} 
In the feedforward process, joint states from the \OURS exoskeleton are mapped to control targets for the follower arm. We group the first three joints $q_{\text{shoulder}}^{\text{\ exo}} = [q_1^{\text{\ exo}}, \ q_2^{\text{\ exo}}, \ q_3^{\text{\ exo}}]^T$ into a virtual spherical joint. Its orientation, represented using a rotation matrix, is computed with forward kinematics:
$\mathbf{R}_{\text{shoulder}}^{\text{\ exo}} = f_{\text{FK}}(q_{\text{shoulder}}^{\text{\ exo}})$, where $\mathbf{R}_{\text{shoulder}}^{\text{\ exo}} \in SO(3)$. 
The target joint values for the follower arm's shoulder are subsequently solved via inverse kinematics: $q_{\text{shoulder}}^{\text{\ robot}} = f_{\text{IK}}(\mathbf{R}_{\text{shoulder}}^{\text{\ exo}})$. Because the elbow is a 1-DoF revolute joint, its position is directly mirrored: $q_{\text{elbow}}^{\text{\ robot}} = q_{\text{elbow}}^{\text{\ exo}}$. The virtual spherical wrist orientation is retargeted in the same way, using the wrist sub-manipulator's corresponding joints.



For velocity feedforward, joint velocity targets are mapped using the respective sub-manipulator Jacobians $J\in\mathbb{R}^{3 \times n}$ for ball-joint rotation. Taking the shoulder as an example, the virtual angular velocity of the shoulder ball joint $\omega_{\text{shoulder}}$ is obtained via: $\omega_{\text{shoulder}} = J_{\text{shoulder}}^{\text{\ exo}} \ \dot{q}_{\text{shoulder}}^{\text{\ exo}}$. The commanded joint velocities for the follower robot are then computed using the pseudo-inverse of the follower's shoulder Jacobian: $\dot{q}_{\text{shoulder}}^{\text{\ robot}} = \big( J_{\text{shoulder}}^{\text{\ robot}} \big)^{\dagger} \omega_{\text{shoulder}}$, where $J^\dagger=(J^TJ)^{-1}J^T$. For a non-redundant, 3-DoF shoulder sub-manipulator ($J\in\mathbb{R}^{3\times3}$), the pseudo-inverse can be simplified to the standard matrix inverse $J^{\dagger} = J^{-1}$.

By decoupling the arm into isolated sub-manipulators, our retargeting architecture avoids the kinematic singularities that typically plague 6D end-effector task-space control~\cite{khatib1987operational}.
Because computations for each sub-manipulator remain isolated, their respective Jacobian inverses do not interfere with each other.
This isolation significantly enhances tracking stability and provides predictable, intuitive control near complex kinematic boundaries in constrained space.




\textbf{Haptic Torque Feedback.} 
Given the measured joint torques of the follower shoulder $\tau_{\text{shoulder}}^{\text{\ robot}}$, the spatial rotational moment $F_{\text{shoulder}}$ is solved using the transpose of the dynamically consistent Jacobian inverse~\cite{khatib1987operational}:
\begin{equation*}
    F_{\text{shoulder}} = \big( \ \bar{J \ }_{\text{shoulder}}^{\text{\ robot}} \ \big)^T \tau_{\text{shoulder}}^{\text{\ robot}},
\end{equation*}
where $J \in\mathbb{R}^{3 \times n}$ is the shoulder Jacobian for ball-joint rotation, $\bar{J \ } = A^{-1}J^T(JA^{-1}J^T)^{-1}$ is its dynamically consistent inverse, and $A$ denotes the joint-space inertia matrix of the follower shoulder. For a non-redundant, 3-DoF shoulder, $J\in\mathbb{R}^{3\times3}$, the dynamically consistent inverse simplifies to the standard matrix inverse, $\bar{J \ } = J^{-1}$. The exoskeleton shoulder feedback torque is then computed via:
\begin{equation*}
    \tau_{\text{shoulder}}^{\text{\ exo}} = \big( \ {J \ }_{\text{shoulder}}^{\text{\ exo}} \ \big) ^T F_{\text{shoulder}}.
\end{equation*}
This formulation is applied symmetrically to the wrist, while the elbow torque is directly mapped: $\tau_{\text{elbow}}^{\text{\ exo}} = \tau_{\text{elbow}}^{\text{\ robot}}$. Finally, to maximize the interaction torque transparency, we compute the exoskeleton's whole-arm commanded torque with Coriolis, centrifugal, gravity, and friction compensation. See Appendix~\ref{appendix:exo_control} for full implementation details about the control and retargeting algorithms.


\mysubsection{Portable Mobile Manipulation Teleoperation}


To minimize cost for \OURS, we use IMU for exoskeleton orientation detection and mobile base control. We compute the gravity-compensation torque based on the exoskeleton's orientation relative to gravity. Simultaneously, we linearly map the 3-dimensional Euler angle representation of the IMU to the mobile base velocity vector $[\dot{x}, \dot{y}, \dot{\theta}]$ to control the movement of the mobile base. We choose IMU ($\sim$\$9) for its low cost, while more expensive alternatives like VR headsets ($\sim$\$300) could serve as a drop-in replacement. Our mobile robot base is an upgraded version of Tidybot++ \cite{holmberg2000development,wu2024tidybotopensourceholonomicmobile}, and achieves better compliance and more accurate odometry through a lower reduction ratio and backlash, while remaining kinematically holonomic. For details on breakdown costs and how we build an entire in-house mobile manipulator, see Appendix~\ref{appendix:mobile_manipulator}.

\mysubsection{Learning Autonomous Whole-body Active Compliant Policies}



We use Action Chunking with Transformers (ACT)~\cite{Zhaoetal2023} for learning whole-body compliant robot policies. Details on the model parameters and training process can be found in Appendix~\ref{appendix:learning}. We use the same input-output parameterization for image and robot joint proprioception as the original ACT algorithm. However, in order to incorporate the torque modality for learning active compliant behaviors, we process torque data in the same way as joint positions, and then append their embeddings together to further extend the image embeddings from the ResNet18~\cite{he2016deep} backbone. The final embeddings are then fed into the transformer encoder and decoder networks~\cite{vaswani2017attention} to get the desired joint positions. The desired positions encode joint compliance information to determine the robotic arm's output torque.

\section{Experiments}

\begin{figure}[h]
    \vspace{-4mm}
    \centering
    \includegraphics[width=\linewidth]{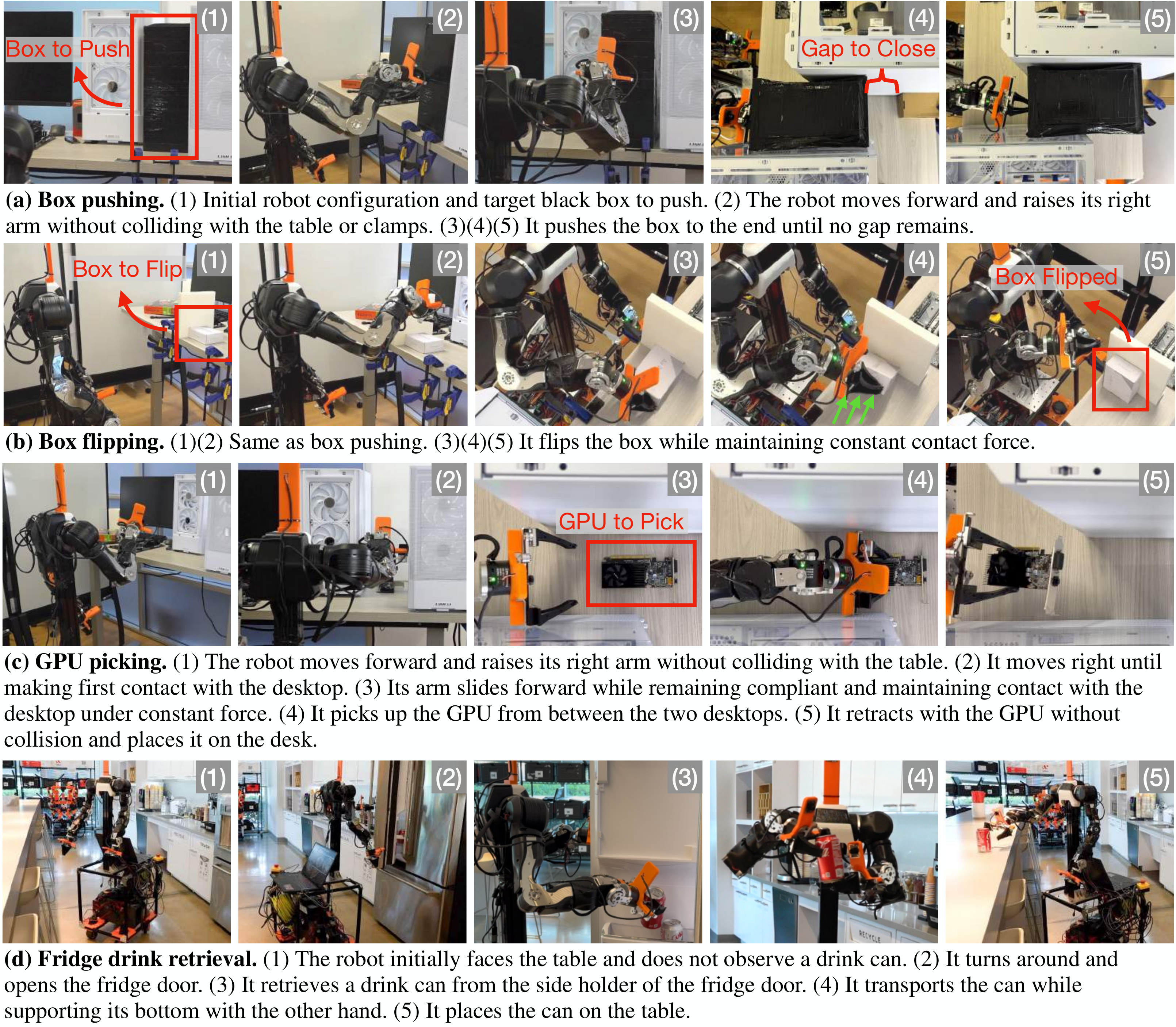}
    \caption{\footnotesize Task description. We provide autonomous task execution videos of all four tasks at \url{https://ume-exo.github.io}.}
    \label{fig:tasks_description}
    \vspace{-2mm}
\end{figure}

We study \OURS's ability to collect data that enables successful learning of autonomous, compliant, bimanual, and whole-body robot policies (video demonstrations can be found at \url{https://ume-exo.github.io}). 
To study \OURS's performance on mobile manipulation tasks, we built an in-house dual-arm mobile manipulator (see Appendix~\ref{appendix:mobile_manipulator} for details). Our tasks are specifically designed to demonstrate the contributions of the three capabilities introduced in Sec.~\ref{bullet}, briefly reiterated here: \hyperref[item:cap1]{Cap1}. real-time haptic torque feedback, \hyperref[item:cap2]{Cap2}. universal whole-arm configuration capture, and \hyperref[item:cap3]{Cap3}. portability for mobile manipulation.

In particular, we study four tasks: (1) box pushing, (2) box flipping, (3) GPU picking, and (4) fridge drink retrieval. Task definitions can be found in their respective subsections, and snapshots of the robot autonomously performing each task are shown in Fig.~\ref{fig:tasks_description}. These tasks span a range of difficulties, summarized below.

\begin{itemize}[leftmargin=*, itemsep=0pt, topsep=0pt]
    \item \label{item:ch1} \textbf{Ch1. Highly force-mediated}. Tasks such as box flipping and box pushing require the learned policy to effectively and consistently regulate the interaction forces applied to the environment. This challenge demonstrates the importance of \hyperref[item:cap1]{Cap1}.  
        
    \item \label{item:ch2} \textbf{Ch2. Tightly space-constrained}. Tasks such as GPU picking are designed to operate in highly constrained spaces where contact is unavoidable~\citep{thomasson2025slim,brouwer2025gentle,thomasson2022going}, so the policy must constantly react to force/torque information to achieve active compliant behavior. The policy also needs to implicitly learn collision avoidance from whole-arm-configuration demonstrations. This demonstrates the importance of \hyperref[item:cap1]{Cap1} and \hyperref[item:cap2]{Cap2}. 
        
    \item \label{item:ch3} \textbf{Ch3. Requiring mobile manipulation}. Tasks such as fridge drink retrieval require long-horizon whole-body control. In fact, all four tasks require a certain level of base movement. This challenge demonstrates the importance of \hyperref[item:cap3]{Cap3}. 
\end{itemize}


We compare \OURS (ours) against the two baselines below across all four tasks over 20 trials.

\begin{itemize}[leftmargin=5mm, itemsep=0pt, topsep=0pt]

    \item \textit{No-torque.} To ablate the importance of real-time interaction torque, we train a baseline policy on the same dataset without including the torque modality to compute the proprioception embedding.
    
    \item \textit{UMI.} To ablate the importance of the whole-arm configuration, we trained a baseline policy using only the end-effector 6D pose in the same dataset as the ground-truth actions. We trained with the same network input-output parameterization as UMI~\cite{Chietal2024}, namely: input image, relative end-effector pose, and inter-end-effector pose, and output the next chunk of relative end-effector pose, without including torque data. We implemented a whole-body IK solver to convert end-effector poses into joint positions.

\end{itemize}

\begin{wraptable}{r}{0.5\textwidth}
    \vspace{-10mm}
    \addtolength{\tabcolsep}{-3pt}
    \footnotesize
    \centering
    \begin{tabular}{c|cccc}
    \toprule
    & \makecell{Box \\ Pushing} & \makecell{Box \\ Flipping} & \makecell{GPU \\ Picking} & \makecell{Fridge Drink \\ Retrieval} \\
    \midrule
    \textit{No-torque} & $0.50$ & $0$ & $0.75$ & $0.90$ \\
    \textit{UMI} & $0.40$ & 0 & $0$ & $0$ \\
    \textit{UME (Ours)} & \textbf{0.90} & \textbf{0.85}& \textbf{0.95} & \textbf{0.95} \\
    \bottomrule
    \end{tabular}
    \vspace{2mm}
    \caption{\footnotesize \OURS and baselines task success rate.}
    \label{tab:results}
    \vspace{-4mm}
\end{wraptable}


\mysubsection{Autonomous Policy: Box Pushing}
\textbf{\underline{Task}} \ Push a target box into the space between two desktop PCs until it reaches the very end, leaving no gap between the box and any hidden objects behind it (Fig.~\ref{fig:tasks_description}(a)). This task is inspired by how humans clean tabletops by shuffling objects all the way against the wall.


\textbf{\underline{Challenges}} \ This task is highly force-mediated (\hyperref[item:ch1]{Ch1}). The hidden objects behind the large target box are not visually observable, so the robot must rely entirely on torque information to determine whether any space remains. The number, location, and size of gaps created by these hidden objects are arbitrary and unknown, so simply memorizing a stopping location based on visual input or joint values will not work. 
In addition, the robot must move its base in order to push the box all the way to the end (\hyperref[item:ch3]{Ch3}). 

\textbf{\underline{Performance}} \ We collected 26 demonstrations,
and \OURS achieves a success rate of 0.9 (18/20).

\begin{figure}[t]
    \vspace{-2mm}
    \centering
    \includegraphics[width=\linewidth]{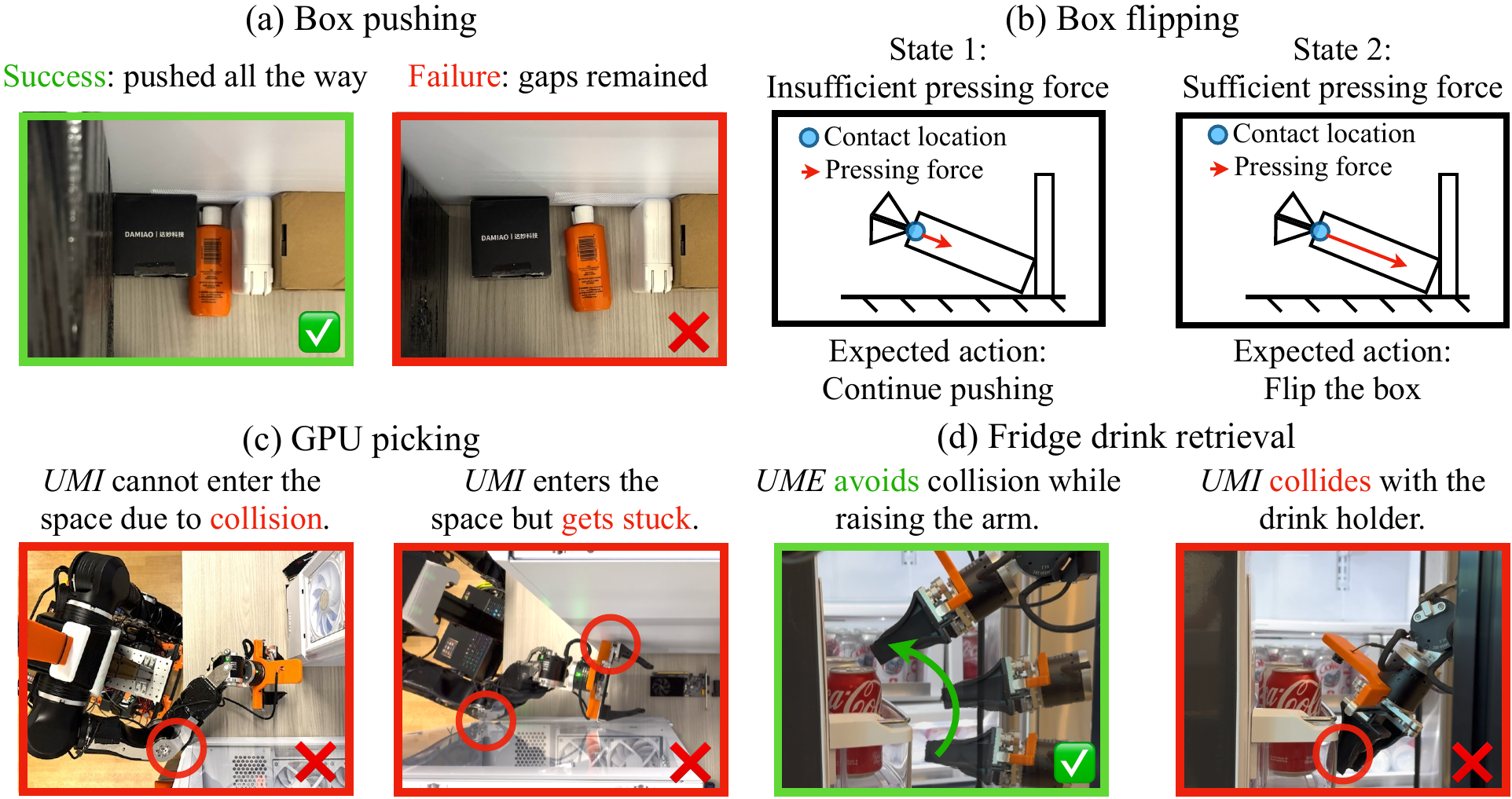}
    \caption{\footnotesize Ablations and representative failure cases of baselines.}
    \label{fig:ablations}
    \vspace{-3mm}
\end{figure}

\textbf{\underline{Ablations}} \ Removing the torque modality severely reduces the \textit{No-torque} baseline success rate to 0.5. \textit{No-torque} often fails to push the target box to close all gaps between the hidden objects (Fig.~\ref{fig:ablations}(a)). 
During data collection, the human demonstrator relies entirely on instantaneous torque feedback, continuing to apply pressure until a high torque indicates that no gaps remain. Without access to the torque information, \textit{No-torque} prematurely stops midway while pushing the box. Moreover, without torque feedback, it is very easy for the demonstrator to overdrive the box and crash any fragile objects behind it. This task thus highlights the critical importance of real-time haptic feedback both for human demonstration and for policy learning. 

\textit{UMI} struggles for similar reasons due to the lack of torque information, and it also learns to stop midway. It can sometimes succeed when there is only a small gap behind the box to close. In addition, \textit{UMI} also suffers from collisions with the environment, leaving its performance at 0.4, slightly worse than \textit{No-torque}.

\mysubsection{Autonomous Policy: Box Flipping\label{section:box_flipping}}

\textbf{\underline{Task}} \ Push a box against a vertically fixed surface to flip the box upright (Fig.~\ref{fig:tasks_description}(b)).


\textbf{\underline{Challenges}} \ This task is highly force-mediated (\hyperref[item:ch1]{Ch1}). To successfully flip the box upright, the robot must maintain a constant pressing force against the box to prevent it from falling. In particular, the robot must distinguish between two critical states that appear visually identical but require different actions based on joint torque readings (Fig.~\ref{fig:ablations}(b)). In the first state, the gripper is in contact with the box, but the applied pressing force is insufficient, so the expected action is to continue increasing the end-effector force. In the second state, sufficient contact force (consequently, friction) has been established, and the expected action is to initiate the flipping motion. Successful completion of the task requires interleaving between these two state-action pairs, which can only be reliably distinguished through torque sensing. In addition, similar to box pushing, we clutter the tabletop with realistic computer parts to make collision avoidance necessary (\hyperref[item:ch2]{Ch2}). At the beginning of the task, the robot starts very close to the table and must raise its arm from below the tabletop while avoiding contact with the table and surrounding objects.

\textbf{\underline{Performance}} \ With 40 demonstrations in the training dataset, \OURS achieves a success rate of 0.85 (17/20). 

\textbf{\underline{Ablations}} \ Without torque information, both \textit{No-torque} and \textit{UMI} are unable to complete the task (0 success rate), as neither can distinguish between the two aforementioned states. In the majority of failure cases, the arm attempts a flipping motion without actually flipping the box because the pressing force is insufficient to generate enough friction for the box to move. In other cases, the box slips midway through the flipping motion. See our website for videos of representative failure cases.


\mysubsection{Autonomous Policy: GPU Picking}

\textbf{\underline{Task}} \ The robot must reach between two desktops to pick up a GPU card located at the back of the constrained space, and retrieve it to a location outside the enclosure (Fig.~\ref{fig:tasks_description}(c)).

\textbf{\underline{Challenges}} \ This task happens in an extremely constrained space (\hyperref[item:ch2]{Ch2}), with only $\sim$5mm clearance after the gripper is inserted between the desktops. It also requires base movement (\hyperref[item:ch3]{Ch3}) to reach the end of the space.

To complete this task, the robot must first move its arm from underneath the desk to above the tabletop while avoiding collisions, and then fully straighten the arm, which is the only configuration that can reach the back of the constrained space. Even a slight bend in the arm would cause it to collide with one of the desktops. This again demonstrates the importance of recording and learning from whole-arm configurations (\hyperref[item:cap2]{Cap2}). 

In such a tightly constrained environment, contact is almost unavoidable. As a result, the policy must leverage torque feedback to compliantly slide along one side of the desktop while gradually reaching toward the back. It must continuously regulate the contact force between the gripper and the desktop (\hyperref[item:ch1]{Ch1}) to prevent excessive scratching of the glass surface. In the video on our website, the gripper can even be heard gliding along the desktop's glass surface. Finally, at the end of the retrieval phase, the gripper must carry the GPU out of the constrained space and place it on the desk while still avoiding collisions with the opposite desktop. 

\textbf{\underline{Performance}} \ With 42 demonstrations, \OURS achieves a high success rate of 0.95 (19/20). 

\textbf{\underline{Ablations}} \ By learning from whole-arm configurations and torque information, \textit{UME} learns to embrace contact and continuously regulate the contact force with the desktop while sliding along the glass surface, successfully handling all the challenges. Without the torque modality, \textit{No-torque} still learns to straighten the arm, but it sometimes applies excessive force to the side surface of the desktop, preventing further forward motion or scratching the desktop surface. In contrast, \textit{UMI} is unable to complete the task at all. When learning only from end-effector poses, it is extremely difficult for the policy to avoid collisions. As shown in Fig.~\ref{fig:ablations}(c), \textit{UMI} consistently bends its arm either before or after entering the constrained space, leading to collisions.

\mysubsection{Autonomous Policy: Fridge Drink Retrieval}

\textbf{\underline{Task}} \ The robot initially faces the table and checks whether a drink can is present. If no drink is detected, the robot turns around, opens the fridge door, picks up a drink from the side holder of the door, transports it while using the other gripper to support the bottom of the can, and places it on the table (Fig.~\ref{fig:tasks_description}(d)). If the drink is removed afterward, the robot repeats the same procedure, ensuring that a drink can is always available on the table. Note that the can is unopened and full, not empty.

\textbf{\underline{Challenges}} \ This task requires full coordination between mobile movement and bimanual control (\hyperref[item:ch3]{Ch3}), with one arm holding the fridge door while the other grasps the drink. This task also has the longest horizon among all evaluated tasks (lasting $\sim$1 min 30 sec). Moreover, the task is force-mediated (\hyperref[item:ch1]{Ch1}), as torque feedback is important for successfully opening and holding the heavy fridge door.

\textbf{\underline{Performance}} \ The training dataset contains 157 demonstrations, and our policy achieves a success rate of 0.95 (19/20). The policy is highly robust to external disturbances. First, if the drink is removed after being placed on the table, the robot always turns around and retrieves another drink. Second, if the drink is taken away during transport, the robot drops its arms midway through the episode, returns to the fridge, and resumes the task from there. See our website for video demonstrations. 

\textbf{\underline{Ablations}} \ Since the fridge drink retrieval task is long-horizon, we further break down the success rate into multiple stages (see Appendix \ref{appendix:fridge_stage_wise} Table~\ref{tab:fridge_stage_wise}). \textit{No-torque} also achieves a high overall success rate of 0.90, although it occasionally fails to apply sufficient torque to open the door or collides with the environment when retrieving. While \textit{UMI} also fails once to open the fridge door due to insufficient gripping force, it never successfully grasps and lifts the drink out of the side holder. This is primarily caused by collisions with the environment, either from the wrist-mounted camera or the gripper itself (shown in Fig.~\ref{fig:ablations}(d)). In comparison, our policy consistently maneuvers around the holder to avoid collisions and forms a stable grasp on the drink. 

\begin{wrapfigure}{r}{0.25\textwidth}
    \vspace{-8mm}
    \centering
    \includegraphics[width=\linewidth]{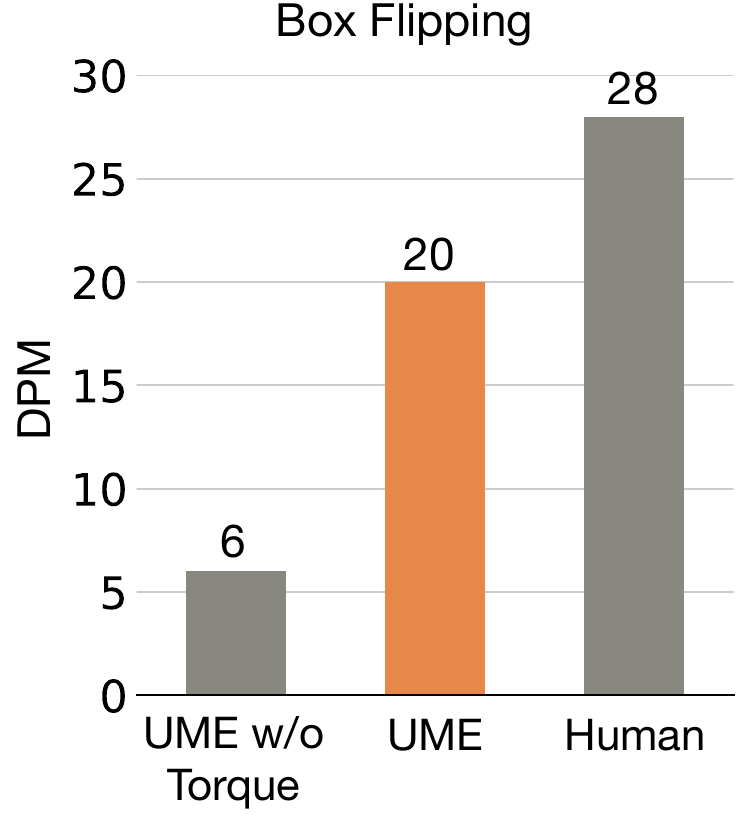}
    \caption{\footnotesize Demonstrations per minute (DPM) over multiple users.}
    \label{fig:user_study}
    \vspace{-4mm}
\end{wrapfigure}

\mysubsection{Data Collection Throughput and Universal Teleoperation}

\textbf{Data Collection Throughput.} \OURS's haptic torque feedback improves user experience, resulting in higher data collection throughput. To demonstrate this, we average the number of demonstrations per minute (DPM) over different users using 3 different methods: 1) \OURS without torque feedback, 2) \OURS, and 3) human demonstrations (see Fig.~\ref{fig:user_study_demo} in Appendix~\ref{appendix:user_study}). We measure the data collection throughput on the box flipping task (results shown in Fig.~\ref{fig:user_study}). Note that the reset time between each episode is also included in the one-minute period. On this highly force-mediated task, \OURS achieves a 3.3$\times$ improvement over the baseline without torque, operating at 71\% of human speed.


\textbf{Universal Teleoperation.} To demonstrate \OURS's cross-embodiment generality, we use it to teleoperate bimanual X-ARMs~\cite{xarm} in the real world and Franka~\cite{franka} arms in simulation. We successfully teleoperate the X-ARMs to pick up a block, transfer it between arms, and place it into a cup (see Fig.~\ref{fig:other_arms} in Appendix~\ref{appendix:other_arms}). Demonstrations are available on our website: \url{https://ume-exo.github.io}.

\section{Conclusion}
\label{sec:conclusion}

We present \OURSFULL (\OURS), a low-cost, lightweight, and portable exoskeleton that captures whole-arm configuration while providing real-time haptic torque feedback for teleoperation. Through careful kinematic design---including coaxial actuator configurations and backdrivable high-torque actuators---\OURS enables intuitive control across a range of robot arms with different DoFs. With our universal retargeting algorithm and an onboard IMU, \OURS supports robot-agnostic whole-body mobile manipulation. These capabilities enable robust end-to-end learning of highly force-mediated, visually occluded, space-constrained, and long-horizon manipulation tasks directly from human demonstrations.

\section{Limitations and Future Work}

While \OURS demonstrates strong performance across a range of tasks, several limitations remain. First, the exoskeleton links are built with PLA plastic, resulting in higher weight and lower payload compared to more expensive alternatives. Future work could develop a lighter, production-grade version that remains low-cost. Second, currently the links have fixed dimensions. A new version with adjustable link lengths could improve accessibility and ergonomics. Finally, due to delays in Franka arm delivery, we only demonstrate Franka teleoperation in simulation. We plan to evaluate the system on real Franka arms once they arrive, and expect similar or improved performance given Franka's more accurate joint torque sensing.




\clearpage
\acknowledgments{
This work is supported by Ant Group.

We would like to thank Hao Li and Tian-Ao (Teo) Ren from the Stanford BDML Lab for insightful discussions on the mechanical design of \OURS. We are grateful to Yifan Hou, Zhanyi Sun, and Professor Shuran Song from the Stanford REAL Lab for valuable discussions on autonomous robot experiment design and future applications. We are grateful to Mikael Jorda for feedback on the design of the retargeting algorithm.

We would also like to thank Xiangpeng Miao from Damiao Technology for actuator suggestions and support, Heng Le from Ufactory X-ARM for assistance with the robotic arms, and Haoxing Guo from Hexfellow Robotics for support with the power caster module.
}


\bibliography{references}  

\clearpage

\appendix
{\Large \textbf{Appendix}}

\section{Extended Related Work}
\label{sec:related_work}

\textbf{Teleoperated data collection.} 
Teleoperation~\citep{darvish2023teleoperation,peternel2022after} has long been used for demonstration data collection, and since the introduction of ALOHA~\cite{Zhaoetal2023}, it has emerged as one of the dominant paradigms.
Many follow-up works have further improved different aspects of the original system, including ALOHA 2~\cite{aldaco2024aloha}, Mobile ALOHA~\cite{fu2024mobile}, FACTR~\cite{liu2025factr}, and GELLO~\cite{Wu2024GELLO}. In \textit{ALOHA-style teleoperation}, the user drives the end-effector of a leader arm that shares a similar morphology with the follower arm. However, controlling the whole-arm configuration is not intuitive unless the user uses both hands to physically constrain a single leader arm.

%
%
%

Compared to ALOHA-style teleoperation, \textit{exoskeleton-based teleoperation} provides a more intuitive way to control the entire arm configuration. While upper-limb exoskeletons have long been popular in rehabilitation robotics~\citep{zimmermann2019anyexo,nef2009armin,klamroth2014armin,kim2017harmony,zimmermann2023anyexo,mao2012carex,bergamasco2016human,bergamasco2016human}, researchers have recently begun exploring their use for collecting robot demonstrations. 
HOMIE~\citep{ben2025homie} is an isomorphic exoskeleton that commands a humanoid robot directly in joint space 
but the mobility of the humanoid is controlled using an additional foot pedal, unlike our embedded IMU. 
AirExo~\citep{fang2024airexo} and AirExo-2~\citep{fang2025airexo} can also directly command the follower arm in joint space, but they are not fully isomorphic, requiring additional calibration steps or a specialized adapter. Despite adopting an exoskeleton form factor, ACE~\citep{yang2024ace} records only wrist poses and finger joint values. 
Another line of work in teleimpedance~\citep{peternel2022after} builds impedance-command interfaces to directly control the stiffness of the follower arm. However, as far as we know, the majority of the aforementioned works do not provide direct joint-level haptic torque feedback to the operator. Their focus is primarily on motion capture and position-based teleoperation rather than bidirectional torque interaction.



\textbf{In-the-wild data collection.}
Another popular paradigm for collecting robot demonstrations is in-the-wild data collection that does not require a robot. One of the main categories is handheld devices~\cite{Chietal2024,Xuetal2025DexUMI,choi2026inthewildcompliantmanipulationumift,xu2026hommi,etukuru2024rum,seo2025legato,shafiullah2023bringing,young2020vime,song2020grasping}. Handheld devices are typically designed with end-effectors identical to those of the robot. By equipping both the handheld devices and the robot with wrist-mounted cameras, these methods leverage in-hand observations to ensure visual consistency. In addition to handheld devices, some exoskeletons~\cite{fang2024airexo,fang2025airexo} can also be used to collect demonstrations directly rather than teleoperation. In-the-wild data collection naturally enables real-time haptic feedback to the operator through direct physical interaction with the environment. However, because these methods typically record only end-effector poses, policies deployed in highly constrained spaces must rely on external inverse kinematics solvers for collision handling. 

\textbf{Learning active compliant behaviors.} Active compliance in robot manipulation can be achieved through different paradigms. One common approach is to combine a high-level position-based policy with a low-level compliance controller, such as admittance control~\cite{maples1986experiments} or impedance control~\cite{hogan1984impedance}.
Building on this, recent work has explored learning the compliance parameters of such controllers from demonstrations. 
\citet{hou2024acp} introduce Adaptive Compliance Policy (ACP), which augments a diffusion policy to additionally predict stiffness parameters, and uses a fixed admittance controller at the low level. 
In another work, \citet{xu2026compliant} incorporate force/torque feedback into policy learning, but still relies on higher-level motion or residual action outputs combined with compliant controllers. Alternatively, compliance can be achieved by learning an end-to-end torque policy, where the policy directly maps observations to torque commands, bypassing any explicit compliance controller~\citep{kalakrishnan2011learning,lee2019making}. 
While this approach offers greater flexibility, it requires a torque-controllable platform and places the full burden of stable, safe force control on the learned policy.

\section{Extended Analysis on Fridge Drink Retrieval \label{appendix:fridge_stage_wise}}
\begin{table}[h]
    \vspace{-2mm}
    \addtolength{\tabcolsep}{-3pt}
    \footnotesize
    \centering
    \begin{tabular}{ccccccc}
    \toprule
    & \makecell{Turn Around} & \makecell{Open fridge} & \makecell{Take out drink} & \makecell{Close fridge} & \makecell{Transport drink} & \makecell{Place drink} \\
    \midrule
    \textit{No-torque} & \textbf{1.00} & 0.95 & 0.90 & 0.90 & 0.90 & 0.90 \\
    \textit{UMI} & \textbf{1.00} & 0.95 & 0.00 & 0.00 & 0.00 & 0.00\\
    \textit{UME (Ours)} & \textbf{1.00} & \textbf{1.00} & \textbf{0.95} & \textbf{0.95} & \textbf{0.95} & \textbf{0.95} \\
    \bottomrule
    \end{tabular}
    \vspace{2mm}
    \caption{\footnotesize Fridge drink retrieval stage-wise success rate.}
    \label{tab:fridge_stage_wise}
    \vspace{-3mm}
\end{table}

\section{Extended Details on Universal Retargeting}
\label{appendix:exo_control}
\begin{figure}[h]
    \centering
    \includegraphics[width=\linewidth]{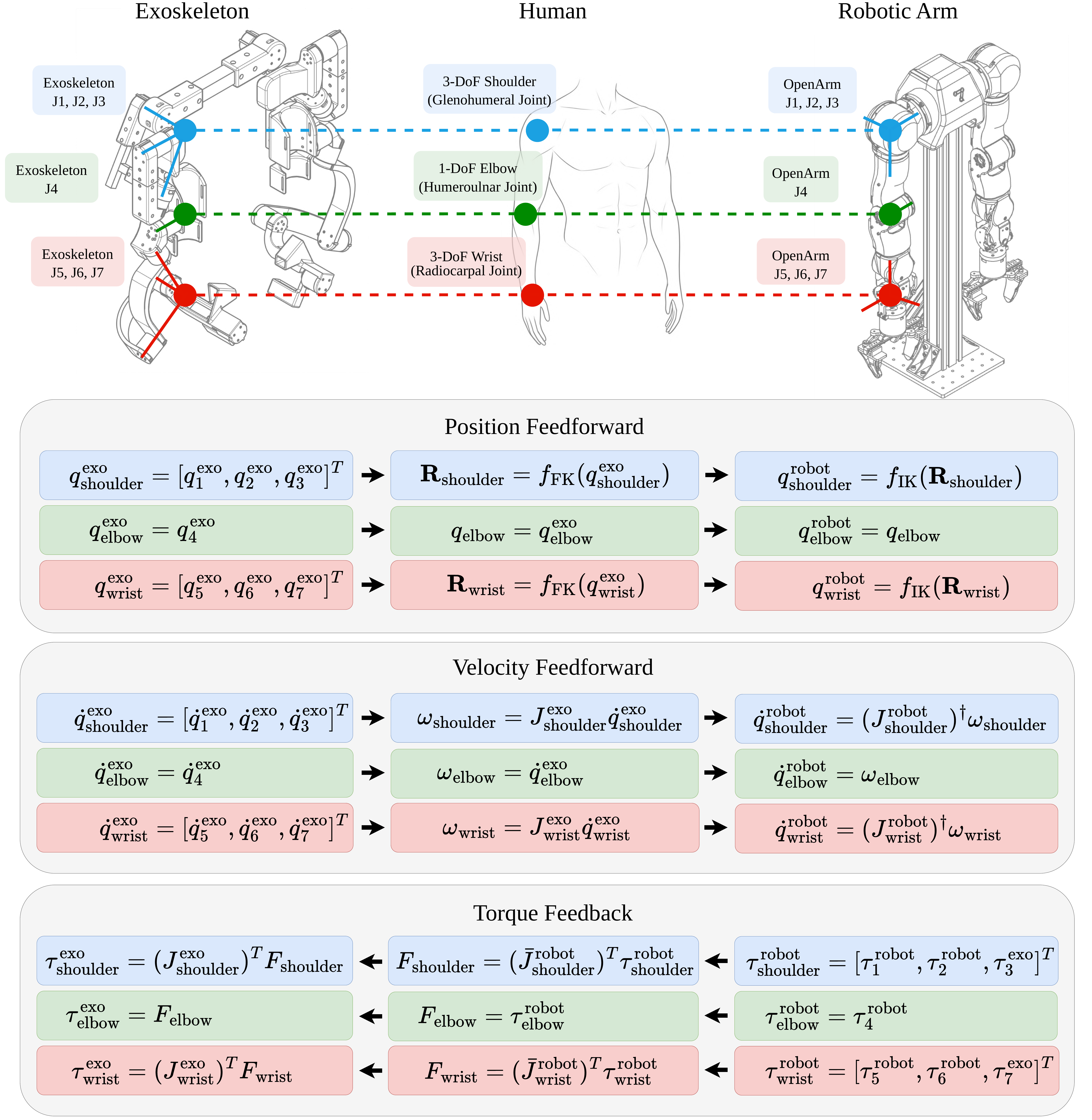}
    \caption{\OURS retargeting algorithm for every sub-manipulator. }
    \label{fig:universal_retargeting_full}
\end{figure}

In this section, we present the complete computation underlying our universal retargeting algorithm. Fig.~\ref{fig:universal_retargeting_full} shows the full retargeting process to compute position and velocity feedforward and torque feedback for all sub-manipulators. 

\subsection{Exoskeleton Commanded Torque}
After we independently obtain the feedback torque for the shoulder, elbow and wrist sub-manipulators, namely, $\tau_{\text{shoulder}}^{\text{\ exo}}$, $\tau_{\text{elbow}}^{\text{\ exo}}$, and $\tau_{\text{wrist}}^{\text{\ exo}}$, there is one more step to compute the exoskeleton’s final whole-arm commanded torque with Coriolis, centrifugal, gravity, and friction compensation, shown below:

\begin{equation*}
    \tau_{\text{command}}^{\text{\ exo}} = c(q^{\text{\ exo}}, \  \dot{q}^{\text{\ exo}}) + g(q^{\text{\ exo}}) + f(\dot{q}^{\text{\ exo}}) + \tau_{\text{feedback}}^{\text{\ exo}}
\end{equation*}

where $c(q^{\text{\ exo}}, \dot{q}^{\text{\ exo}})$ is the Coriolis and centrifugal compensation torque, $g(q^{\text{\ exo}})$ is the gravity compensation torque, $f(\dot{q}^{\text{\ exo}})$ is the friction compensation torque, and $\tau_{\text{feedback}}^{\text{\ exo}} = \left[\tau_{\text{shoulder}}^{\text{\ exo}},\ \tau_{\text{elbow}}^{\text{\ exo}},\ \tau_{\text{wrist}}^{\text{\ exo}} \right]$ is the collated feedback torque from all sub-manipulators.

\subsection{Teleoperating X-ARM and Franka}
\label{appendix:other_arms}
\begin{figure}[h]
    \centering
    \includegraphics[width=\linewidth]{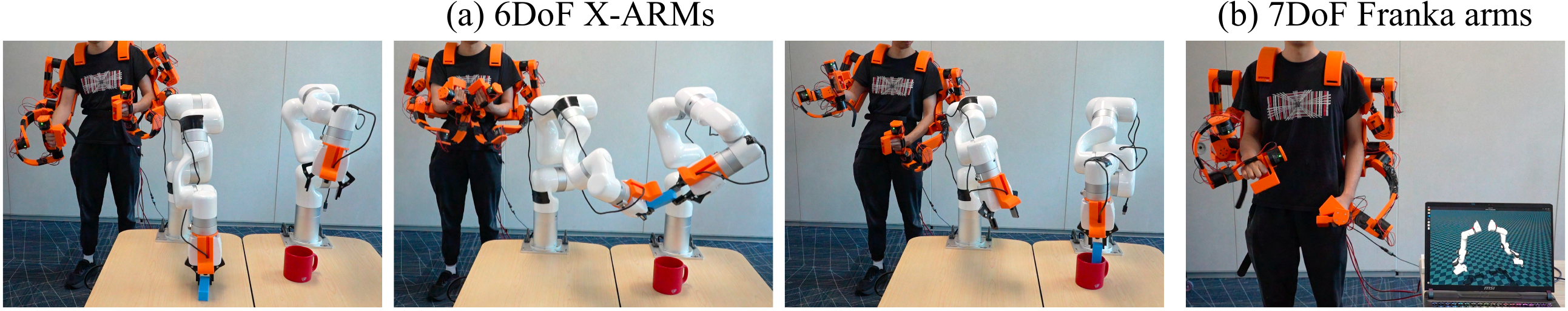}
    \caption{\footnotesize \OURS teleoperates other robotic arms. }
    \label{fig:other_arms}
    \vspace{-3mm}
\end{figure}

To further demonstrate the universal retargeting algorithm, we use \OURS to teleoperate bimanual 6-DoF X-ARMs and 7-DoF Franka arms (shown in Fig.~\ref{fig:other_arms}). The operator can control the X-ARMs to pick up a block, hand it over to the other arm, and place it into a cup. Since physical hardware is unavailable for the Franka arms, we demonstrate their teleoperation using the MuJoCo simulator~\citep{todorov2012mujoco}.

\section{Extended Details on the Learning Algorithm}
\label{appendix:learning}

Table~\ref{tab:learning_parameters} shows parameters for training our ACT models. All models are trained for 40k gradient steps on one NVIDIA RTX 4090 GPU for 8 hours.

\begin{table}[h]
    \addtolength{\tabcolsep}{-3pt}
    \footnotesize
    \centering
    \begin{tabular}{cc}
    \toprule
    Parameter & Value \\
    \midrule
    batch size & 200 \\
    image backbone architecture & ResNet18 \\
    image backbone learning rate & $1\times10^{-5}$ \\
    transformer encoder layers & 4 \\
    transformer decoder layers & 7 \\
    feedforward dimension & 3200 \\
    hidden dimension & 512 \\
    transformer heads & 8 \\
    transformer learning rate & $1\times10^{-4}$ \\
    dropout & 0.1 \\
    \bottomrule
    \end{tabular}
    \vspace{2mm}
    \caption{\footnotesize Parameters for autonomous policy learning.}
    \label{tab:learning_parameters}
    \vspace{-3mm}
\end{table}

\section{Hardware Details}
\label{appendix:hardware}

In this section, we provide extra details on \OURS and our in-house mobile manipulator. 

\subsection{\OURS Motors, Materials and Weight}
\begin{figure}[h]
    \centering
    \includegraphics[width=0.7\linewidth]{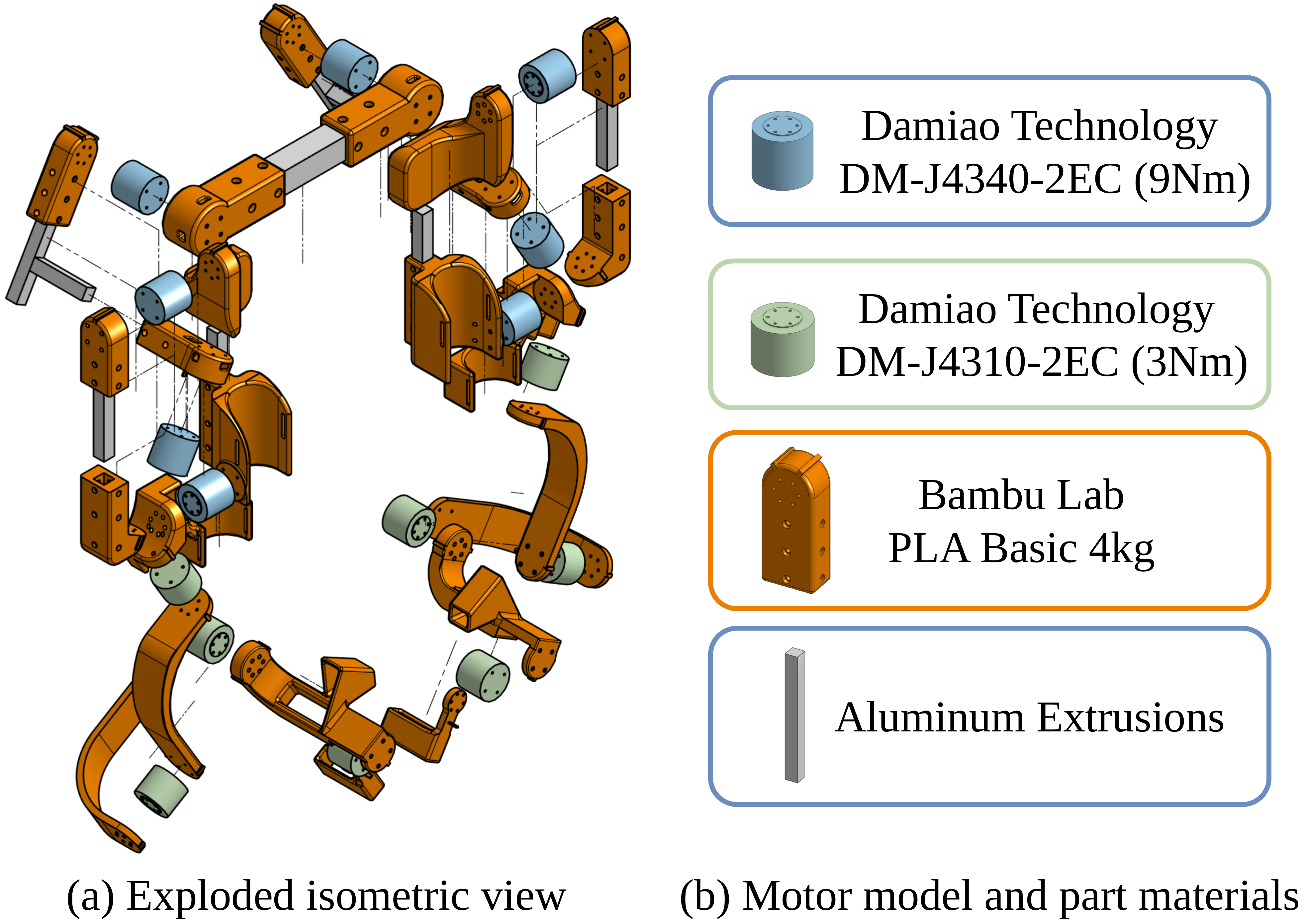}
    \caption{\footnotesize \OURS motors and materials. }
    \label{fig:exploded_view}
\end{figure}

An exploded isometric view of \OURS, along with the motor models and materials used for different components, is provided in Fig.~\ref{fig:exploded_view}. To balance portability, haptic torque transparency, and the torque required for payload support and gravity compensation, we selected the Damiao Technology DM-J4340-2EC as the shoulder actuator because it best satisfies our design requirements. The DM-J4340-2EC provides 9Nm of torque while weighing only 362g with a 40:1 reduction ratio (9Nm, 362g, 40:1). We evaluated several similar actuators but ultimately rejected them due to their trade-offs. For example, the DM-J4340P (9Nm, 375g, 40:1) uses a cross-roller bearing that increases multi-directional stiffness at the cost of higher friction, reducing torque transparency. The DM-J8006 (8Nm, 550g, 6:1), in contrast, achieves a lower reduction ratio at the cost of a significantly heavier stator coil, reducing portability.

Built with PLA plastic structural links, \OURS weighs approximately 12kg, allowing a typical adult to wear it comfortably for up to 2 hours. Using higher-strength and stiffer materials, such as CNC-machined aluminum, could further reduce the weight to approximately 8kg.

\subsection{\OURS Bill of Materials}
\OURS is designed with cost efficiency in mind from day one. The total material cost of building a complete \OURS system is only \$1900, roughly the price of a MacBook. To achieve this low cost, we intentionally select off-the-shelf components, including Damiao~\cite{damiao} joint actuator modules, Bambu Lab~\cite{bambu} PLA 3D printing material~\cite{bambu}, and industry-standard hardware such as M3/M5 screws and 2020 aluminum extrusions. A detailed breakdown of the material selection and associated costs is provided in the bill of materials (Table~\ref{tab:exo_bom}).

\begin{table}[h]
    \addtolength{\tabcolsep}{-3pt}
    \footnotesize
    \centering
    \begin{tabular}{cc}
    \toprule
    Material Name & Price (USD$\times$Count)  \\
    \midrule
    {Damiao Technology DM-J4340-2EC}$\times 8$ & \$$117\times 8$ \\
    {Damiao Technology DM-J4310-2EC}$\times 8$ & \$$88\times 8$ \\
    {IXGNIJ $2020\times20$cm aluminum extrusion} & \$$3.5\times6$\\
    {Bambu Lab PLA Basic}$\times 4$ & \$$13\times 4$ \\
    {Kunhong dual channel CAN-FD to USB adapter} & \$$52\times 1$\\
    {Jesverty adjustable 30V 10A DC power supply} & \$$41\times 1$ \\
    {Gaoyuan XT30 2+2 cable}$\times14$ & \$$2\times14$ \\
    {Yahboom IMU} & \$$9\times1$ \\
    {M3 hex socket screws} & \$$3\times6$ \\
    {M5 hex socket screws} & \$$5\times6$ \\
    {M5 T-nuts} & \$$7\times1$\\
    {Quick-detach belt} & \$$2\times 1$ \\
    \midrule
    Total & \$$1900$ \\
    \bottomrule
    \end{tabular}
    \vspace{2mm}
    \caption{\footnotesize Bill of materials for one \OURS.}
    \label{tab:exo_bom}
    \vspace{-3mm}
\end{table}

\subsection{\OURS Range of Motion}
\label{appendix:range_of_motion}

\begin{figure}[h]
    \centering
    \includegraphics[width=\linewidth]{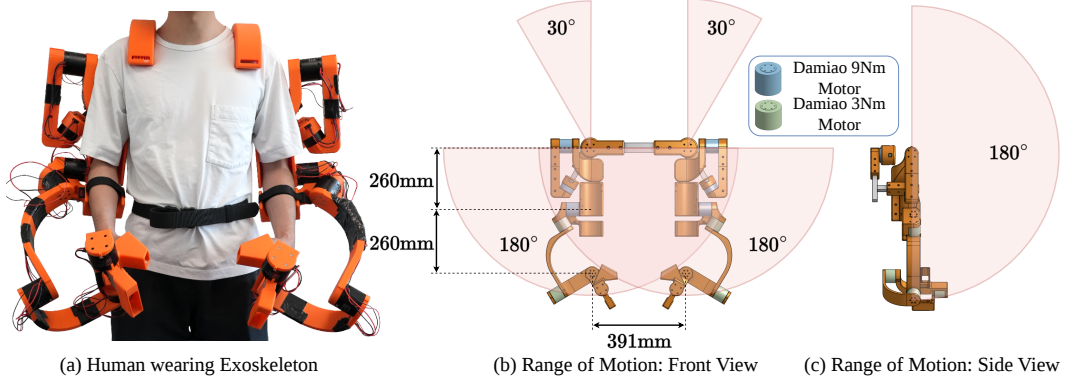}
    \caption{\footnotesize \OURS Range of motion. }
    \label{fig:range_of_motion}
    \vspace{-3mm}
\end{figure}

\OURS captures the human whole-arm configuration with minimal compromise to the range of motion. Unlike other leader–follower teleoperation systems, \OURS preserves nearly the entire wrist range of motion, while only limiting the reachability of certain uncommon shoulder postures. Its range of motion is visualized from both the front and side views in Fig.~\ref{fig:range_of_motion}.

\subsection{In-House Mobile Manipulator Platform}
\label{appendix:mobile_manipulator}
To evaluate \OURS's capability for collecting demonstrations to learn autonomous mobile manipulation policies, we built an in-house mobile manipulator. We discuss several key design choices below.

\begin{figure}[h]
    \centering
    \includegraphics[width=0.7\linewidth]{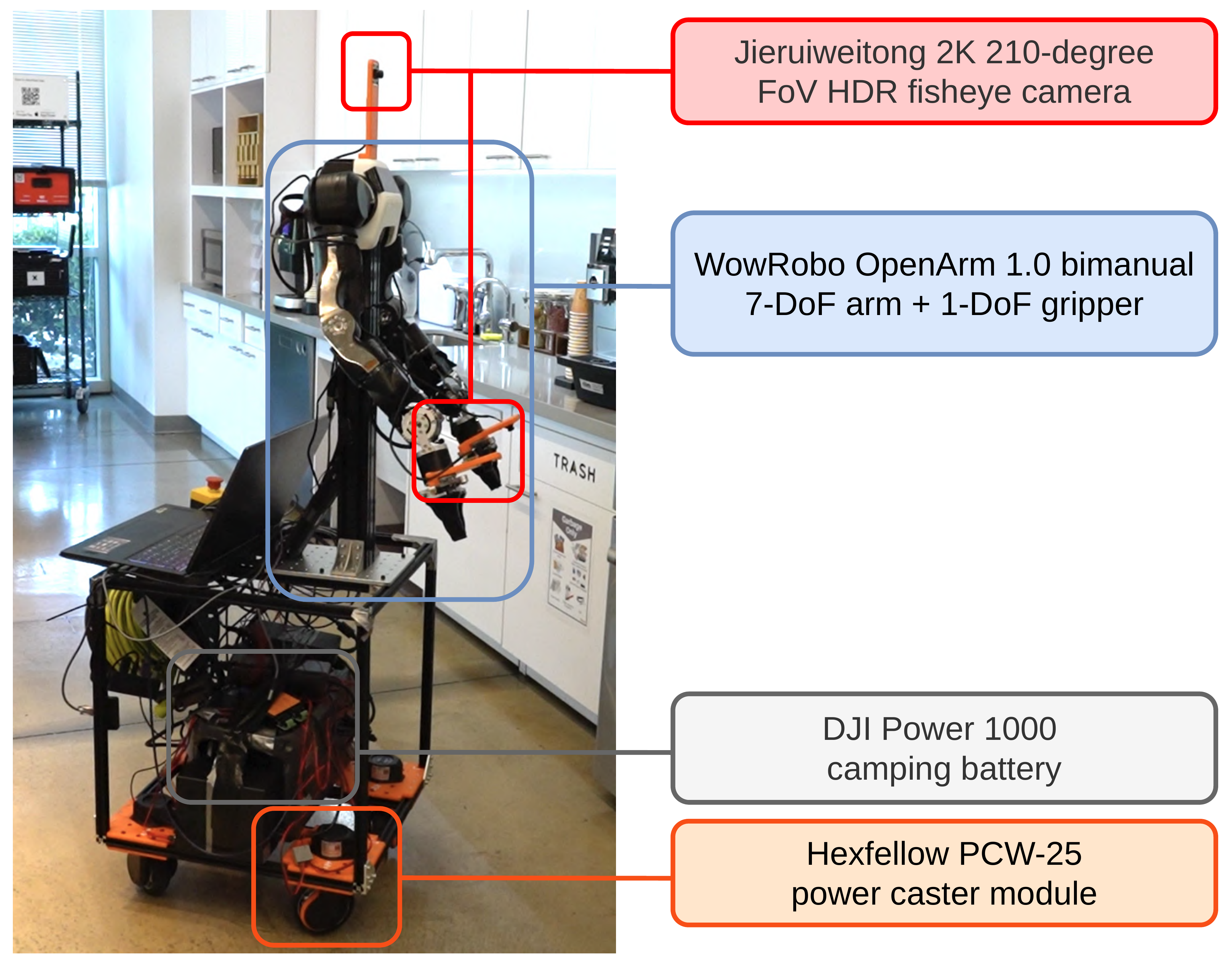}
    \caption{\footnotesize In-house mobile manipulator key components. }
    \label{fig:robot_materials}
    \vspace{-3mm}
\end{figure}

\textbf{Robot Arm.} \ We select the WowRobo~\cite{wowrobo} OpenArm 1.0 bimanual 7-DoF arm due to its high-quality Damiao~\cite{damiao} joint actuators (DM-J8009P, DM-J4340P, DM-J4340, and DM-J4310), well-machined aluminum components, and low overall cost (\$5200 for the complete upper body with two arms).

\textbf{Mobile Base.} \ By using the Hexfellow PCW-25~\cite{hexfellow} (\$900) powered caster module instead of the FRC SDS MK4 swerve module~\cite{sds} (\$300) used in the original TidyBot++~\cite{wu2024tidybotopensourceholonomicmobile} system, our mobile robot base achieves improved compliance and more accurate odometry due to its lower gear ratio and reduced backlash, while remaining fully holonomic. We use the DJI Power 1000~\cite{dji} to supply power to all robot electronics, eliminating the need for a wired wall connection during mobile manipulation. We choose the DJI Power 1000~\cite{dji} over other available options due to its well-designed power delivery system, including surge protection, well-isolated multi-channel AC outputs, and overall build quality.

\textbf{Camera.} \ We use Jieruiweitong~\cite{jieruiweitong} 2K-resolution, 210$^\circ$ FoV (field of view) HDR fisheye cameras on the robot head and wrists. We select these cameras for their low cost (\$37.5), wide FoV, low latency, and strong high-dynamic-range imaging performance.


\subsection{In-House Mobile Manipulator Bill of Materials}

Table~\ref{tab:robot_bom} provides a detailed breakdown of the material selection and cost for our in-house mobile manipulator.

\begin{table}[h]
    \vspace{-2mm}
    \addtolength{\tabcolsep}{-3pt}
    \footnotesize
    \centering
    \begin{tabular}{cc}
    \toprule
    Material Name & price (USD$\times$Count)  \\
    \midrule
    WowRobo OpenArm 1.0 & \$$5200\times1$ \\
    Hexfellow Robotics PCW-25$\times4$ & \$$900\times4$ \\
    DJI Power 1000 camping battery & \$$386\times1$ \\
    Jieruiweitong fisheye camera$\times3$ & \$$37.5\times3$ \\
    Kunhong dual channel CAN-FD to USB adapter & \$$52\times 1$\\
    Jesverty adjustable 60V 5A DC power supply & \$$45\times 1$ \\
    IXGNIJ $2020\times50$cm aluminum extrusion$\times14$ & \$$7.5\times14$\\
    2020 corner bracket $\times16$ & \$$1.6\times16$\\
    M5 T-nuts & \$$7\times1$\\
    \midrule
    Total & \$$9533.1$ \\
    \bottomrule
    \end{tabular}
    \vspace{2mm}
    \caption{Bill of materials for one in-house mobile manipulator.}
    \label{tab:robot_bom}
    \vspace{-3mm}
\end{table}

\section{User Study \label{appendix:user_study}}
\begin{figure}[h]
    \centering
    \includegraphics[width=0.6\linewidth]{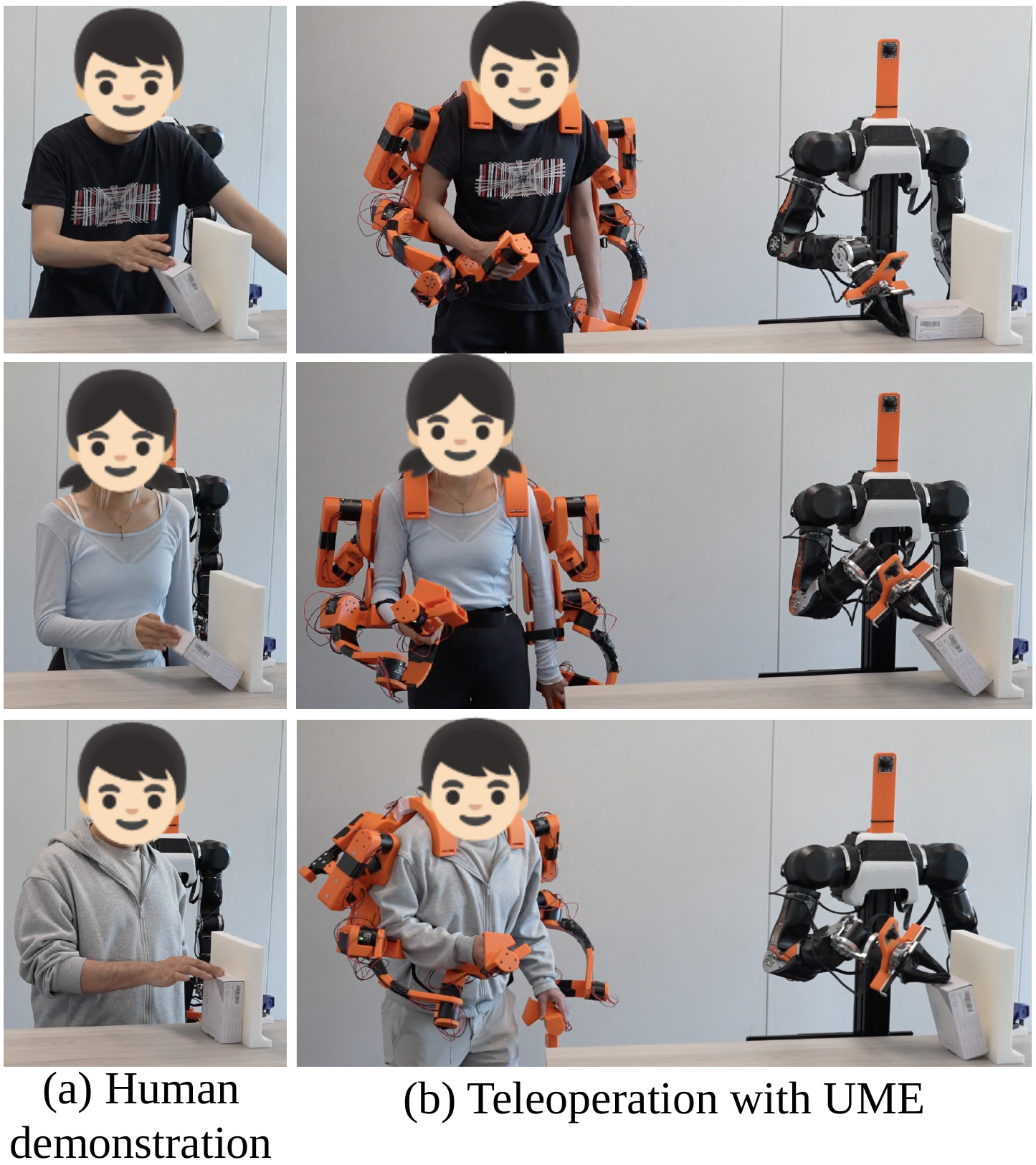}
    \caption{\footnotesize User study on different users.}
    \label{fig:user_study_demo}
\end{figure}

Fig.~\ref{fig:user_study_demo} demonstrates two of the three methods compared in the user study. Human demonstrations provide an upper bound on the performance of \OURS teleoperation.

\end{document}